\pdfoutput=1
    
\documentclass[11pt]{article}

\usepackage[]{acl}

\usepackage{times}
\usepackage{latexsym}
\usepackage{pdfpages}
\usepackage{subcaption}
\usepackage[T1]{fontenc}
\usepackage{rotating}

\usepackage[utf8]{inputenc}

\usepackage{microtype}

\title{Semi-self-supervised Automated ICD Coding} 

\author{Hlynur D. Hlynsson$^{1}$, Steindór Ellertsson$^{2}$, Jón F. Daðason$^{1}$, Emil L. Sigurdsson$^{2, 3, 4}$, Hrafn Loftsson$^{1}$   \\
  $^{1}$ Department of Computer Science, Reykjavik University,
Reykjavik, Iceland \\
  $^{2}$ Primary Health Care Service of the Capital Area, Reykjavik, Iceland \\
  $^{3}$ Department of Family Medicine,
University of Iceland, Reykjavik, Iceland \\
$^{4}$ Development Centre for Primary Health Care in Iceland, Reykjavik, Iceland \\
  \texttt{\{hlynurh, jond19, hrafn\}@ru.is} \\ 
  \texttt{steindor.ellertsson@gmail.com, emilsig@hi.is} \\}

\begin{document}
\maketitle
\begin{abstract}
Clinical Text Notes (CTNs) contain  physicians' reasoning process, written in an unstructured free text format, as they examine and interview patients. 
In recent years, several studies have been published that provide evidence for the utility of machine learning for predicting doctors' diagnoses from CTNs, a task known as ICD coding.
Data annotation is time consuming, particularly when a degree of specialization is needed, as is the case for medical data.
This paper presents a method of augmenting a sparsely annotated dataset of Icelandic CTNs with a machine-learned data imputation in a semi-self-supervised manner. We train a neural network on a small set of annotated CTNs and use it to extract clinical features from a set of un-annotated CTNs. These clinical features consist of answers to about a thousand potential questions that a physician might find the answers to during a consultation with a patient. The features are then used to train a classifier for the diagnosis of certain types of diseases.
We report the results of an evaluation of this data augmentation method over three tiers of information that are available to a physician. Our data augmentation method shows a significant positive effect, which is diminished when an increasing number of clinical features, from the examination of the patient and diagnostics, are made available. We recommend our method for augmenting scarce datasets for systems that take decisions based on clinical features that do not include examinations or tests.
\end{abstract}

\section{Introduction}

When a patient consults a physician, communication is created in the patient's medical records. The physician notes down the patient's signs, symptoms, results of physical examination, the clinical thinking process, and if any diagnostic tests are warranted -- in a free text format known as a Clinical Text Note (CTN). Then, the physician saves the diagnoses, using the International Classification of Diseases (ICD) code, that they made during the consultation. Thus, each CTN contains free text, from which clinical features can be extracted, in addition to the ICD classification code.


Previous work has shown the benefits of training machine learning classifiers on clinical features for automated ICD coding \citep{liang2019evaluation, ellertsson2021artificial, zhang2020bert, pascual2021towards, kaur2021systematic, blanco2021contribution}.  
\citet{ellertsson2021artificial}  hand-annotated features in 800 CTNs and trained a classifier to predict ICD codes for one of four types of primary headache diagnoses. 
\citet{liang2019evaluation} hand-annotated a significantly larger set, i.e. about 6,000 CTNs, for the purpose of training a classifier to predict various types of diseases, i.e. 55 ICD codes in total. 
Additionally, \citeauthor{liang2019evaluation} 
developed a clinical feature extraction model (CFEM), for the purpose of automatically extracting features from the CTNs. 

On its own, the CFEM is beneficial because it could solve the common clinical problem of getting a quick and comprehensive overview of a patient, when meeting a clinician for the first time.  A clinician could search a patient's medical history with a question such as ``Has the patient ever had a colonoscopy?''. The ICD classifiers have, on the other hand, the potential of being integrated into a Clinical Decision Support System (CDSS), where they could, for example, predict if a physician should order an MRI for a patient when presented with a particular symptom, what kind of blood tests are warranted, or any other diagnostic test for that matter.

 Generally, machine learning systems require large quantities of training data \citep{hlynsson2019measuring} and ICD classifiers are no exception. In order to develop a high accuracy ICD classifier, without annotating large amount of CTNs, we experiment with a method of: 1) annotating a small subset of the CTNs with question-answer pairs which are used for training the CFEM, and then 2) use the trained feature extractor to extract clinical features from a larger dataset of CTNs for training the classifier to predict one out of six ICD codes\footnote{The ICD classes were chosen by doctors according to their perceived usefulness. }. We call this method semi-self-supervised because it lies at the intersection of 1) semi-supervised learning, which combines a small amount of labeled data with large amounts of unlabeled data~\citep{van2020survey} and 2) self-supervised learning, which learns to predict missing parts of inputs~\citep{mao2020survey}.


In prior work on ICD coding, classifiers are trained on discharge summaries, after the patient has left the clinic \citep{liang2019evaluation, zhang2020bert, pascual2021towards, kaur2021systematic, blanco2021contribution}. We instead focus on evaluating our model on stages in the primary health care pipeline where the recommendations of machine learning models would be the most effective. We thus introduce a novel three-tiered evaluation system that is designed to mirror the circumstances where ICD classification methods would actually be used and we evaluate our semi-self-supervised data augmentation method on these three tiers: 1) before the patient meets a physician, 2) after the physician performs the patient examination, and 3) after the physician has ordered diagnostic tests.
 
 Our evaluation results show that the data augmentation method has a significant benefit for tier 1, i.e. before the patient meets a physician, but not for the other two.

\section{Related Work}
\label{sec:related}
\citet{liang2019evaluation} frame the  problem of clinical feature extraction from CTNs as a question-answering task. Every clinical feature mentioned in a given CTN is marked, as well as the start and the end of the text span referring to a given clinical feature. A question is saved in the context of the text span, which contains the answer to that specific question. For example, given the text span ``the patient has a fever'', the question ``Does the patient have a fever?'' is saved with a binary value of 1. Out of 1.3 million CTNs from a single institution in China, \citeauthor{liang2019evaluation} annotated about 6,000 CTNs for training a CFEM, based on a Long Short-Term Memory (LSTM) network~\citep{hochreiter1997long} enriched with word embeddings. The feature extractor is trained on a batch of (CTN, question, text span) tuples as input with the goal of optimizing for the text span that contains the corresponding answer to the question in the given CTN. Thereby, the model learns to extract relevant clinical features from the questions put forward in the context of the CTN. \citeauthor{liang2019evaluation} used the CFEM to extract features from the whole set of un-annotated CTNs. The extracted features were then used to train a classifier, based on multi-class logistic regression, to predict an ICD code from a set of 55 codes.

\citet{ellertsson2021artificial} hand-annotated clinical features (in a similar manner as \citeauthor{liang2019evaluation}) in 800 CTNs from a common medical database of all primary care clinics in Iceland. Each CTN had an accompanying ICD code for one of four types of headache diagnoses. The resulting features (text spans) were then used to train a Random Forest classifier, for predicting one of the four possible ICD codes.  Furthermore, they performed a retrospective study where the classifier was shown to outperform general practitioners on the four types of headache diagnostics. 

 In this paper, we expand upon the work of \citeauthor{ellertsson2021artificial} The main difference between our work and theirs can be summarized as follows:
\begin{itemize}
    \item We do not compare our ICD classifier to general practitioners. 
    \item We hand-annotate questions-answers pairs in 2,422 CTNs, which includes a larger number of ICD codes, 42 in total (see Table~\ref{tab:icd_cfem} in the Appendix).
    \item Using the hand-annotated CTNs, we train CFEMs, based on Transformer models~\citep{vaswani2017attention}, for extracting clinical features, and compare them to a couple of LSTM models. These feature extractors are used to extract features from un-annotated CTNS as well as annotated CTNs.
    \item We perform a three-tiered evaluation of our classifiers on six of the ICD codes for pediatric (under 18) patients (see Table~\ref{tab:icd_cpm} in the Appendix). 
\end{itemize}


\begin{table*}[bht!]
\small
\centering
\begin{tabular}{ll|rrrr}
\hline
& & Training Set    & Validation Set  & Test Set & Total\\
\hline
 \rule{0pt}{10pt} Adults & Total size & 1700 & 199  & 220 & 2119 \\
  & Mean Age $\pm$ Std & 45.33 $\pm$ 17.91 & 43.54 $\pm$ 17.86 & 44.24 $\pm$ 17.92 & \\
       & Min Age -- Max Age & 18.01 -- 94.43  & 18.04 -- 86.75  & 18.17 -- 93.72 &  \\ \hline
   \rule{0pt}{10pt} Children & Total size & 237 & 33  & 33 & 303 \\
   & Mean Age $\pm$ Std & 10.01 $\pm$ 5.87  & 10.32 $\pm$ 5.82 & 9.39 $\pm$ 6.24 & \\
       & Min Age -- Max Age & 0.17 -- 17.99  & 0.97 -- 17.85  &  0.21 -- 17.85 &  \\ \hline
\hline
\end{tabular}
\caption{\textbf{Training data split statistics for the clinical feature extraction model.} The adult sets are $63\%$ female and the child sets are $64\%$ female. The different sizes of the adult validation and test sets came by to enforce a constraint of an equal proportion of notes corresponding to each ICD code within each set.}
\label{tab:dataset_split}
\end{table*}

Transformer-based models 
have rapidly become a popular choice for automated ICD coding. These models have been trained on CTNs in a fully end-to-end manner \citep{zhang2020bert, pascual2021towards, kaur2021systematic, blanco2021contribution}. A drawback of this approach is that physicians will often write down their hypothesized diagnoses which injects a serious bias to the data, a problem that our approach, of using one model for clinical feature extraction and another for clinical prediction, circumvents. For example, a fully end-to-end machine learning model might learn to associate the qualitative comment by a physician ``the patient probably has a migraine without aura'' in a patient with a migraine-without-aura ICD code. Our method avoids this by creating a bottleneck of information, where only specific questions are being answered. 

Our approach also opens the door for interpreting the results of the ICD classifier, as the importance of each input feature to the classifier can be visualized, for example by portraying input coefficients in the case of linear models (e.g. logistic regression) or plotting other interpretability metrics, such as SHAP values~\cite{lundberg2017unified}.

\section{Approach}

\subsection{Data and annotation}
\label{subsec:data}

 We use the dataset from the same source as \citet{ellertsson2021artificial}, i.e. from the Primary Health Care Service of the Capital Area (PHCCA) in Iceland. The dataset consists of 1.2 million CTNs, written in Icelandic, from 200 thousand unique patients that were collected in clinical consultations taking place from January 2006 to April 2020. 
 Physicians are instructed not to write anything that can uniquely identify their patients in the notes, but we also used a combination of a parsing system for Icelandic~\citep{thorsteinsson2019wide} as well as a regex command to remove any personally identifiable information, such as names, personal identification numbers and phone numbers.
 This dataset contains CTNs that have an associated ICD code, but consist otherwise of unstructured text from which clinical features can be extracted.

In the same manner as described by \citeauthor{ellertsson2021artificial}, we reduced the full dataset by applying a filter which only keeps notes that contain any word from a medical keyword dictionary. From this reduced dataset, we randomly selected 2,422 notes which were manually annotated by a physician\footnote{The annotator is a white Icelandic male physician in his thirties, specializing in general practice / family medicine.}, resulting in question-answer pairs as described in Section \ref{sec:related}.

As an example annotation, for a CTN containing the text ``the patient is not coughing'', one clinical feature is the pair consisting of the question ``does the patient have a cough?'' and the binary-valued answer ``0'', with the corresponding text span ``not coughing''. Some answers are continuous-valued, such as for the question ``what is the patient's blood pressure?''.

The number of clinical features that we use to train the extraction model to output is 942. There is typically a heavy class imbalance for each feature, where the binary questions have on average a $0.75$ positive answer ratio, with a standard deviation of 0.2. The reason for this sweeping class imbalance is that physicians generally only ask questions that are relevant and with an affirmative answer. 

For our three-tiered classifier evaluation, we define three strict subsets of these features, as described in Section~\ref{sec:threetiered}. Each question is also paired with another binary variable which indicates whether an answer to that question can be found in the CTN or not.


The dataset is split into adults, that are 18 years old or older, and children. Within each age group, $80\%$ of the dataset is allocated for training, $10\%$ for development/validation, and hold out $10\%$ for final testing (see Table~\ref{tab:dataset_split}). The split is stratified to ensure that each set has an equal proportion of sexes and ICD codes.



\subsection{Pre-trained Transformer-based models}
\label{sec:pretraining}
We compared four existing Transformer-based models in our experiments, based on the ELECTRA~\citep{clark2020electra} and RoBERTa~\cite{liu2019roberta} architectures. We evaluated an ELECTRA-small\footnote{\url{https://huggingface.co/jonfd/electra-small-igc-is}. CC-BY-4.0 license.}, ELECTRA-base\footnote{\url{https://huggingface.co/jonfd/electra-base-igc-is}. CC-BY-4.0 license.} and two RoBERTa-base models\footnote{\url{https://huggingface.co/mideind/IceBERT}. AGPL 3.0 license.}$^{,}$\footnote{\url{https://huggingface.co/mideind/IceBERT-igc}. AGPL 3.0 license.} (consisting of 14M, 110M and 125M parameters, respectively). All models have been pre-trained on the Icelandic Gigaword Corpus (IGC)~\citep{steingrimsson2018risamalheild}, which consists of approximately 1.69B tokens from genres such as news articles, parliamentary speeches, novels and blogs. For one of the RoBERTa models, which we refer to as RoBERTa+, the IGC was supplemented with texts obtained from online sources, increasing the size of the pre-training corpus to 2.7B tokens. The RoBERTa models were pre-trained for 225k steps with a batch size of 2k. Otherwise, all models were pre-trained using default settings. The pre-training process and additional training data for the RoBERTa models is described in further detail by~\citet{mideind2022icebert}.


\subsection{LSTM architectures}
\label{subsec:lstmarc}
For a baseline comparison, we created two LSTM models. 
The first one (LSTM 1) tokenizes and trains the embeddings from scratch, whereas the second one (LSTM 2) pre-processes the inputs with GloVe \citep{pennington2014glove} embeddings.
\subsubsection{LSTM 1}
The model splits up the tokenized input into question and content parts. The content part gets a 256-dimensional embedding and the question gets a 32-dimensional embedding. Each embedding is then passed to its own, uniquely parameterized two-layer bi-directional LSTM model, where each layer has 256 units. 

The outputs from those two parts are then concatenated and used to 1) train a set of dense networks, where one is tasked with predicting whether an answer to the question can be found in the text and, if yes, the other dense network predicts the probability of the answer being affirmative (in the case of binary questions), and 2) predict the start and end indices of the tokens that mark the span of the answer in the context part.

\subsubsection{LSTM 2}

LSTM 2 has the same architecture as LSTM 1, except there is no embedding layer and the inputs have been processed by a pre-trained GloVe model. The GloVe embeddings\footnote{\url{https://github.com/stofnun-arna-magnussonar/ordgreypingar_embeddings/tree/main/GloVe}} where pre-trained on the IGC.

\subsection{Clinical feature extraction models}
\label{subsec:cfem}

We fine-tuned the four Transformer-based models, mentioned in Section \ref{sec:pretraining}, on the hand-annotated data in order to develop a CFEM.  The fine-tuning was carried out in the following manner: starting with the pre-trained transformers weights, the top layer was replaced with a randomly initialized network, and the whole system was then trained end-to-end for question-answering.  We also trained the two LSTM models described in Section \ref{subsec:lstmarc} from scratch for a CFEM comparison.

Each model learns to output the answer span for each question\footnote{If the question is not answered in the CTN, the model outputs an impossible span in the text, which is technically implemented as starting at the $0^{th}$ token (a special ``start'' token) and the $1^{st}$ token of the proper context.} as well as the probability of the answer being affirmative for binary-valued questions. The Transformer-based models were defined and trained using the Transformers~\citep{wolf2019huggingface} and PyTorch libraries~\citep{paszke2019pytorch} and the LSTM models were defined and trained using TensorFlow~\citep{abadi2016tensorflow}.

\begin{figure*}[htp!] \centering{
\includegraphics[scale=0.6]{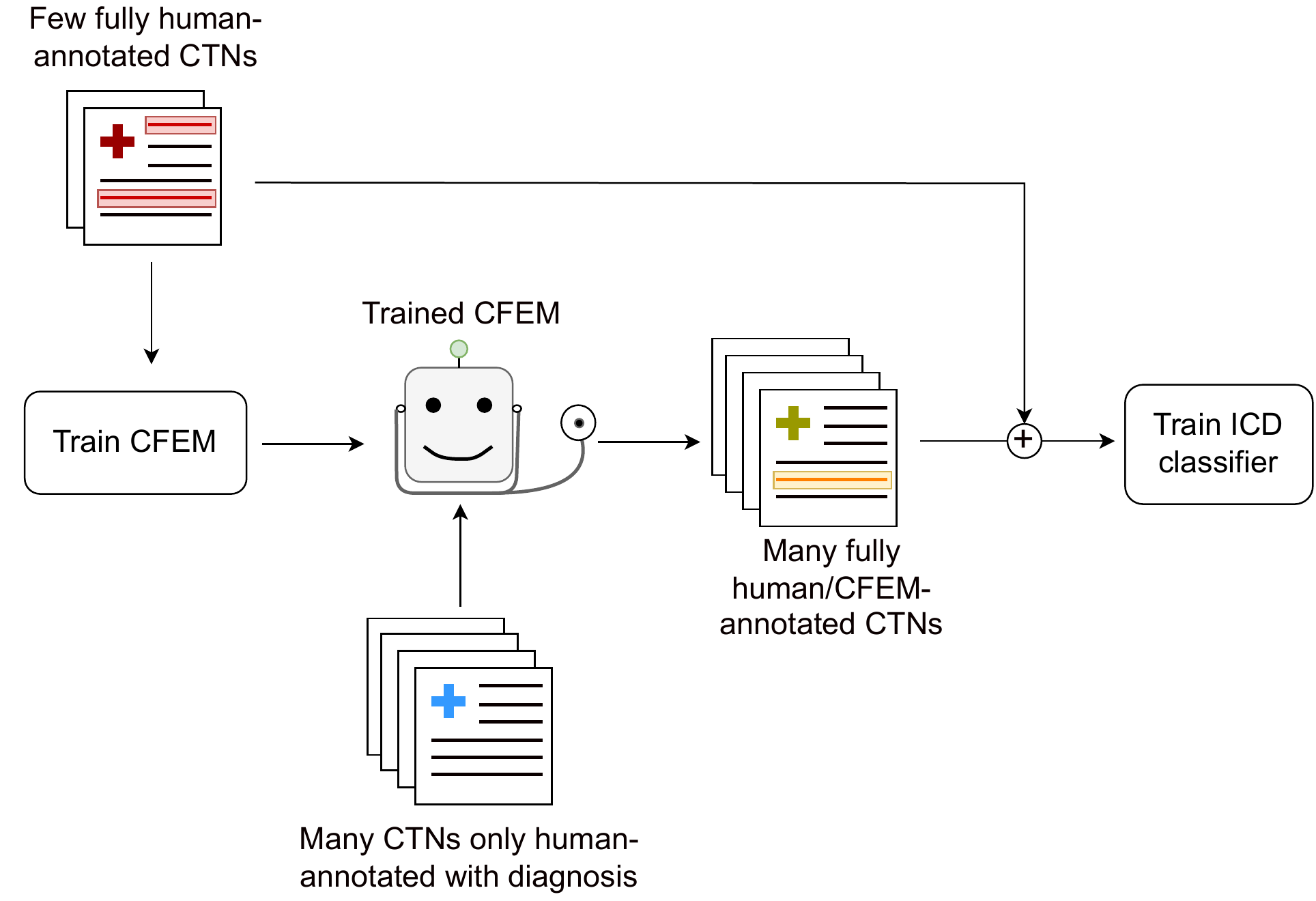}}
\caption{ \textbf{Leveraging a Sparsely Annotated Dataset.} Our clinical feature extraction model learns to mark text spans (clinical features), containing an answer to a set of given clinical questions, from CTNs in which answer spans have been hand-annotated. The feature extractor is then used to extract answer spans -- given the same set of questions -- from a large set of CTNs that have diagnoses (ICD codes), but no marked answer spans. Finally, the extracted answer spans are used to train the ICD classifier. In this way, we make full use of a large set of CTNs that is only partly annotated and combine it with a much smaller set of human-annotated CTNs to learn automated ICD coding.} 
\label{impute}
\end{figure*}

\subsection{Semi-self-supervised learning}
\label{sec:sss}
Once our CFEMs were trained, we saved their outputs over all the CTNs (i.e. 2,422 annotated CTNs used for training and 750 randomly selected un-annotated CTNs) to disk. The outputs define the matrix of independent variables $X$ which is, along with the dependent variable array $y$ of ICD codes, used to train our logistic regression ICD classifier (implemented in scikit-learn~\citep{pedregosa2011scikit}).

CTNs require expertise to interpret, which results in a high cost when labelling medical datasets. This is especially true for AI researchers that are working with a language with much fewer resources than English~\citep{blanco2021contribution}, such as 
Icelandic.

In our project, we have a large collection of CTNs, each of which is marked with a doctor's diagnosis, but does not contain answer spans for the set of questions for our clinical features. We input the un-annotated CTNs to a CFEM, that is trained on a much smaller subset of the data, to take advantage of the supervisory signal offered by the ICD code of each un-annotated CTN. This step keeps the interpretable clinical features and removes potential bias from the CTNs. 
This set of CTNs with imputed clinical feature values is then combined with our ``gold standard'' set of annotated CTNs, and both are used for training the ICD classifier (see Figure~\ref{impute}).

\subsection{Three-tiered evaluation}
\label{sec:threetiered}
To simulate the different stages of a physician's evaluation of a patient in real clinical circumstances, we limit the number of features that are available to the classifier at each stage:

\begin{itemize}
    \item \textbf{Tier 1}: Before a patient meets with a physician. This includes the patient's main complaint, history, symptoms, and vital signs (420 features). 
    \item \textbf{Tier 2:} After the patient has been examined by a physician (582 features).
    \item \textbf{Tier 3:} After results from diagnostics are available (608 features).
\end{itemize}

The full list of features is provided in the Appendix: 
Table~\ref{tier01f} and Table~\ref{tier0f2} for tier 1, which are features that the patient could self-report. Tables~\ref{tier1f} and \ref{tier2f} show the features  for tiers 2 and 3, respectively. After tiers 2 and 3, decisions need to be taken regarding what further tests need to be ordered, for example imaging. 
 
 Note that our system could fit into a triage context at tier 1. The patient could fill out an online questionnaire and get recommendations depending on the results, for example, to go to the emergency room, to go the general physician, or maybe just rest at home with a set of self-care instructions. 

\section{Results and Discussion}

\subsection{Clinical feature extraction model training}

The CFEMs were trained over three epochs on the subset of hand-annotated CTNs (see Table~\ref{tab:dataset_split}). For the ELECTRA-base and RoBERTa-base transformers, each epoch took approximately eight hours on Cloud TPU v3 with eight cores, and half that for ELECTRA-small. The training took approximately three hours for each epoch for the LSTMs. 

The RoBERTa+ model, which is pre-trained on the largest corpus, achieves the best results for all three metrics that we monitor (see Table~\ref{tab:cfem_stats}): a span-based $F_{1}$-score, to evaluate the question-answering portion of the models, and the Matthews correlation coefficient (MCC) \citep{matthews1975comparison, chicco2020advantages} for the binary-valued clinical features (Binary MCC) and for predicting whether the question is answered in the text (Impossible MCC). 

We chose the MCC metric because it is appropriate for imbalanced data \citep{chicco2017ten}  (see discussion of our data in  Section~\ref{subsec:data}) and it offers a suitable combination of the four confusion matrix metrics: true positives, true negatives, false positives and false negatives.

Note in Table~\ref{tab:cfem_stats} that the high $F_{1}$-scores are due to the fact that most questions were correctly predicted to be not answered in any given context. This could be due to the fact that the 15.8 GB corpus, which was used to train RoBERTa+, contains 33 MBs of medical texts. Although this is not a large proportion, it could be enough for the model to have learned transferable representations of medical vocabulary. 

To our surprise, the ELECTRA-base model was outperformed by RoBERTa (both are trained on equal-sized corpora), even though ELECTRA has, previously, been shown to outperform RoBERTa on question-answering tasks~\citep{clark2020electra}. 

The LSTM variation whose inputs were not pre-processed by a pre-trained GloVe model (LSTM 1) performed better according to the MCC metrics (but slightly worse according to the $F_{1}$-score) than the other (LSTM 2). We hypothesize that it is due to the fact that the pre-trained embeddings are not trained with any tokenization, but rather on whole words. The free-text style of doctor's notes can include words or abbreviations that are not defined for the GloVe embeddings.

\begin{table}
\small
\centering
\begin{tabular}{lrrr}
& $F_{1}$ & Binary MCC & Imp. MCC \\[0.1cm] \cline{2-4} 
\multicolumn{1}{l|}{RoBERTa+}     & \textbf{0.993} &  \textbf{0.846} &  \textbf{0.872}\rule{0pt}{10pt} \\[0.1cm]
\multicolumn{1}{l|}{RoBERTa}      & 0.991 & 0.780 & 0.823\\[0.1cm]

\multicolumn{1}{l|}{ELECTRA-base}       & 0.987 &  0.656 &  0.729\\[0.1cm]
\multicolumn{1}{l|}{ELECTRA-small}       & 0.982& 0.553 & 0.650 \\[0.1cm]
\multicolumn{1}{l|}{LSTM 1}       & 0.975& 0.331     & 0.327 \\[0.1cm]
\multicolumn{1}{l|}{LSTM 2}       & 0.979 & 0.313 & 0.257 \\[0.1cm]
\end{tabular}
\caption{\textbf{Feature extraction model evaluation results.} Question-answering metrics and evaluation results for each clinical feature extraction model on the test set.}
\label{tab:cfem_stats}
\end{table}

\subsection{ICD classifier training}

\subsubsection{Transformer vs. LSTM}

After training and evaluating the CFEMs, we validated the data augmentation scheme described in Section~\ref{sec:sss}. We used the best-performing models from each category, RoBERTa+ and LSTM 1, to extract the clinical features from the children's notes\footnote{Due to time constraints, our evaluation of the
data augmentation method is limited to only using
the children CTNs.}. These features, along with their associated ICD codes, were then used to train the classifier. 

Table \ref{tab:lstmttdetails2} shows the diagnostic metrics of the classifier for tier 3 depending on the feature extractor. Using RoBERTa+ yielded a higher weighted average for all diagnostic metrics compared to LSTM 1.

\begin{table*}[!h]
\begin{tabular}{l|llll|llll}
                 & \textit{RoBERTa+} &  &  &  & \textit{LSTM 1} &  &  &   \\
Condition                 & $F_{1}$-score & MCC & TPR & TNR & $F_{1}$-score & MCC & TPR & TNR  \\ \hline
Migraine without aura     & 0.40      & 0.36   & 0.33   & 0.97        & 0.00 & 0.00& 0.00& 1.00 \\
Migraine with aura        & 0.67      & 0.70   & 0.50   & 1.00        & 0.40 & 0.36& 0.33& 0.97 \\
Tension-type headache     & 0.94      & 0.89   & 1.00   & 0.88        & 0.86 & 0.73& 1.00& 0.71 \\
Otitis media, unspecified & 0.00      & 0.00   & 0.00   & 1.00        & 0.57 & 0.60& 1.00& 0.90 \\
Bacterial pneumonia       & 0.86      & 0.83   & 1.00   & 0.93        & 0.75 & 0.75& 0.60& 1.00 \\
Acute bronchitis          & 1.00      & 1.00   & 1.00   & 1.00        & 0.33 & 0.29& 0.25& 0.97 \\ \hline
Weighted average & \textbf{0.81}  & \textbf{0.78} &  \textbf{0.85}  & \textbf{0.85}      & 0.64 & 0.56 & 0.70 & 0.70 
\end{tabular}
\caption{\textbf{Detailed ICD classification metrics.} Per-class metrics for clinical diagnosis prediction when a logistic regression classifier is trained on features extracted from CTNs by either our RoBERTa+ transformer or the baseline LSTM 1 model. MCC is the Matthews correlation coefficient, TPR is the true positive rate and TNR is the true negative rate.}
\label{tab:lstmttdetails2}
\end{table*}

\subsubsection{Qualitative analysis}

To verify that the relationship between our features and the outputs of our models matches our clinical intuition, we 
use SHAP (Shapley additive explanation) values \citep{shapley:book1952} to show the impact of each feature in the prediction of our logistic regression classifier, trained on the features in tier 3 extracted by RoBERTa+. 

\begin{figure*}[!h]
\centering
\begin{subfigure}{.9\textwidth}
  \centering
  \includegraphics[width=.99\linewidth]{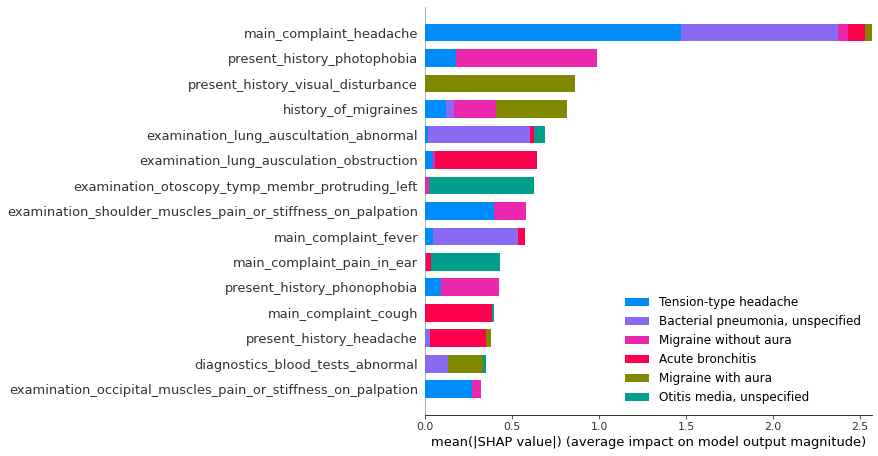}
\end{subfigure}%
\caption{\textbf{Feature importance plot.} The features are scored by their SHAP values. The size of the colored bar in each feature's row indicates the contribution of that feature to predicting the disease with the corresponding color. }
\label{fig:shapvals}
\end{figure*}

The feature importance plot is shown in Figure~\ref{fig:shapvals}.
We see, for example, that the top four features are  headache-related features and  contribute to classifying a CTN as Tension-type headache, migraine with- and without aura. The two top features after that involve the doctor doing a physical examination of the patient's lung and contribute to predicting whether the patient has pneumonia or bronchitis. The sixth most impactful feature is then the result of an examination of the patient's ear, the result of which contributes to the diagnosis of Otitis media (a disease of the middle ear).

\subsubsection{Data augmentation experiment}

In the next set of  experiments, we investigated the effect of augmenting a data set consisting of 303 human-labeled childrens's CTNs with a varying number of machine-labeled children's CTNs for the purpose of training an ICD classifier.

\begin{figure*}[!h]
\centering
\begin{subfigure}{.9\textwidth}
  \centering
  \includegraphics[width=.99\linewidth]{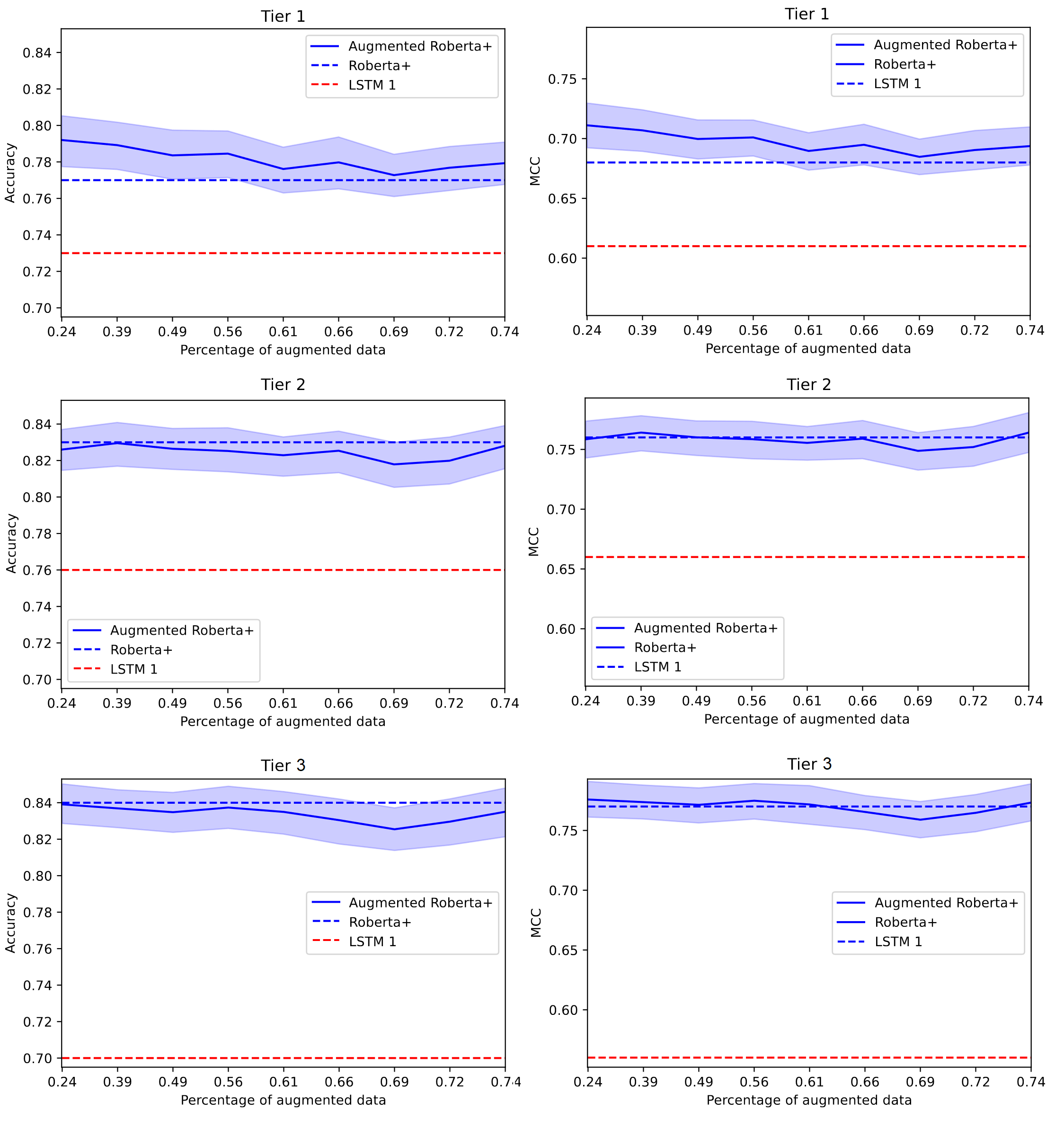}
\end{subfigure}%
\caption{\textbf{Data Augmentation Results.} Each classifier is trained on fixed set of hand-annotated clinical features, in addition to a varying number of features automatically extracted by the RoBERTa+ model, i.e. machine-labeled features. There are 237 hand-annotated CTNs in each training set and each step along the x-axis adds 75 machine-labeled CTNs. Each point in the augmented curves shows the cross-validated metrics (accuracy in the left column and MCC in the right column) averaged over 20 random subsets of machine-labeled points that are added to the training set and the error band (the colored area around the Augmented Roberta+) signifies the $95\%$ confidence intervals. The dashed lines indicate the performance of the classifiers trained only on hand-annotated data.}
\label{fig:data_augm}
\end{figure*}

We trained logistic regression classifiers using 5-fold cross-validation over the whole children set. Each classifier had L1 regularization with the inverse regularization parameter of $C=0.2$, which was found to give good classification performance in early tests. We chose not to do hyper-parameter tuning as the scope of this project is not to get the best possible classifier in this context, but rather investigate the data augmentation and the three-tiered evaluation. The results are shown in Figure~\ref{fig:data_augm}.

There is a clear benefit for using the data augmentation method in tier 1, but it looks rather harmful for tiers 2 and 3. We hypothesize that this is due to the fact that the classifiers place a high importance on the outcome of examination (tier 2) and test (tier 3) related features, making the classifiers more sensitive to prediction errors for these feature.

\section{Conclusions and Future Work}

Our results show that training a CFEM on a small annotated subset of CTNs and use it to extract features from a larger, un-annotated dataset can increase the performance of an ICD classifier. However, the effect is only positive and significant in the context before a patient has been examined by the physician.

A future line of work is to further validate different classifiers by performing prospective studies which allow us to get insight into how the classifier performs in real clinical situations. This can be done by integrating the classifier into a CDSS, where a patient can log into a secure portal, at home or at a medical institution, and answer targeted questions regarding their symptoms. The CDSS could build a list of differential diagnoses, recommend further diagnostics based on the patients symptoms, and then write out the CTN for the clinician. This does not disturb the clinical workflow, saves time for medical staff and potentially allows a much more detailed history taking, compared to the often time constrained clinician. This is important in all outpatient care, both public and private, since this kind of system has the potential to save money, increase the effectiveness and revenue for private clinics without losing the quality of care. 

\section*{Acknowledgements}
This work was funded by the Icelandic Strategic
Research and Development Programme for Language Technology 
and with Cloud TPUs from Google's TPU Research Cloud (TRC).
\clearpage
\bibliography{anthology,custom}
\bibliographystyle{acl_natbib}

\appendix

\section{Appendix}
\label{sec:appendix}


\begin{table}[h!]
\tiny
\begin{tabular}{ll}
\hline
ICD code & Description \\ 
G43.0  &  Migraine without aura \\
G43.1  &  Migraine with aura \\
G44.0  &  Cluster headaches and other trigeminal autonomic cephalgias \\
G44.2  &  Tension-type headache \\
G44.4  &  Drug-induced headache, not elsewhere classified \\
G45.9  &  Transient cerebral ischemic attack, unspecified \\
H66.0  &  Acute suppurative otitis media \\
H66.9  &  Otitis media, unspecified \\
I10  &  Essential (Primary) Hypertension \\
I63.0+  &  Cerebral infarction \\
I63.1  &  Cerebral infarction \\
I63.2+  &  Cerebral infarction due to unsp. occl. or stenosis of precerebral arts. \\
I63.3  &  Cerebral infarction due to thrombosis of cerebral arts. \\
I63.4  &  Cerebral infarction due to embolism of cerebral arteries. \\
I63.5  &  Cerebral infarction due to unsp. occl. or stenosis of cerebral arts. \\
I63.6  &  Cerebral infarction due to cerebral venous thrombosis, nonpyogenic \\
I63.8  &  Other cerebral infarction \\
I63.9  &  Cerebral infarction, unspecified \\
I84  &  Haemorrhoids \\
J00  &  Acute nasopharyngitis [common cold] \\
J01  &  Acute sinusitis \\
J01.0  &  Acute maxillary sinusitis \\
J01.9  &  Acute sinusitis \\
J02.0  &  Streptococcal pharyngitis \\
J03.0  &  Streptococcal tonsillitis \\
J03.9  &  Acute tonsillitis \\
J05.0  &  Acute obstructive laryngitis \\
J10.1  &  Influenza due to other identified influenza virus w/ other resp. manifs.  \\
J11.1  &  Influenza with other resp. manifs., virus not identified \\
J12.9  &  Viral pneumonia, unspecified \\
J15  &  Bacterial pneumonia, not elsewhere classified  \\
J15.7  &  Pneumonia due to Mycoplasma pneumoniae \\
J15.8  &  Pneumonia due to other specified bacteria \\
J15.9  &  Bacterial pneumonia, unspecified \\
J20.9  &  Acute bronchitis \\
J44.1  &  Chronic obstructive pulmonary disease with (acute) exacerbation \\
J44.9  &  Chronic obstructive pulmonary disease, unspecified \\
J45.0  &  Predominantly allergic asthma \\
J45.9  &  Asthma, unspecified  \\
M54.1+  &  Radiculopathy \\
M54.5+  &  Low back pain  \\
S83.2  &  Tear of meniscus, current injury \\
\hline
\end{tabular}
\caption{ICD codes associated with notes used during training of the clinical feature extraction model.}
\label{tab:icd_cfem}
\end{table}

\begin{table}[h!]
\normalsize
\begin{tabular}{ll}
\hline
ICD code & Description \\ 
G43.0  &  Migraine without aura \\
G43.1  &  Migraine with aura \\
G44.2  &  Tension-type headache \\
H66.9  &  Otitis media, unspecified \\
J15.9  &  Bacterial pneumonia, unspecified \\
J20.9  &  Acute bronchitis \\
\hline
\end{tabular}
\caption{ICD codes associated with notes using during classifier training.}
\label{tab:icd_cpm}
\end{table}


\begin{sidewaystable}
\tiny
\centering
\begin{tabular}{llllllllll}
\hline
History of migraines & History smoking & History smoking package years v & History of wiplash & History of alcoholism & History is regularly active & History of bells palsy \\
History of stroke & History of hypertension & History of migraines with descr & History active use alcohol mode & History active substance abuse & History of accident motor vehic & History of cigarette smoking \\
History of head trauma & History smoking active time & History smoking packages per da & History known allergy & History of cluster headache & History of depression & History of anxiety \\
History allergy pollen & History of fibromyalgia & History of headaches & History allergy penicillin & History of osteoarthritis & History of epilepsy & History of copd \\
History of pulmonary cancer & History of ischemic heart disea & History of diabetes mellitus 2 & History of polio & History of hyperlipidemia & History of lupus & History of asthma \\
History of sinusitis & History of diabetes mellitus & History of hypothyrosis & History of palpitations & History of adhd & History of lower back disc prot & History of atrial fib flutter \\
History known medical allergy & History of chrons ds & History of dvt & History of bipolar disease & History allergy sulfa & History tonsillectomy & History appendectomy \\
History hepatitis c & History of prescription drug ab & History of streptococcal pharyn & History of sleep apnea & History of pad & History of lobectomy & History of heart failure \\
History of gastritis & History unilat or bilat catarac & History of pulmonary embolus & History of c section & History of reflux & History of ca mammae & History allergy tramadol \\
History o2 at home & History of heart attack & History of cabg surgery & History of renal cancer & History artificial heart valve & History has pacemaker & History copd gold stage \\
History cardiac catherization d & History of nephrectomy & History of cholecystectomy & History active substance abuse  & History of psoriasis & History cancer prostata & History sick sinus \\
History of gerd & History hiatal hernia & History of recurrent otitis med & History of being prematurely bo & History has one kidney & History of diabetes mellitus 1 & History hysterectomi \\
History of benign prostate hype & History of recurrent pneumonia & History of reumoatoid arthritis & History allergy morphine & History of pulmonary hypertensi & History joint proteses & History smoking time since quit \\
History of kidney stones & History of diverticulitis & History of gout & History allergy ibuprofen & History of substance abuse & History multiple sclerosis & History inactive subtance abuse \\
History active cancer & History of recurrent cystitis & History of chronic renal failur & History of aortic stenosis & History of chest pain & History breast wedge excision & History of glaucoma \\
History of colitis ulcerosa & History of diverticulosis & History of chronic diarrhea & History of compression fracture & History of spinal stenosis & History of dementia & History of heart valve disease \\
History is blind or close to bl & History of parkinsons disease & History has a single lung & History smoking stop year & History of tia & History of eczema & History of iron deficiency \\
History of backpain & History allergy voltaren & History allergy brown band aid & History of pneumonia & History of osteoporosis & History of accident type unkown & Present history nausea \\
Present history tinnitus & Present history shoulder and ba & Present history vomiting & Present history visual disturba & Present history aura & Present history photophobia & Present history recent head tra \\
Present history runny nose & Present history bulbar conjunct & Present history phonophobia & Present history chest pain & Present history dyspnea & Present history fever & Present history limb numbness \\
Present history dizziness & Present history recedes to quie & Present history facial or head  & Present history limb reduced fo & Present history head trauma & Present history ptosis & Present history malaise \\
Present history diplopia & Present history flashing lights & Present history using analgesic & Present history aphasia & Present history nasal congestio & Present history wakes up with s & Present history is hearing chan \\
Present history abdominal pain & Present history feeling unbalan & Present history vertigo & Present history dizzyness on he & Present history syncope & Present history dysphasia & Present history memory problem \\
Present history visual disturba & Present history visual disturba & Present history headache & Present history insomnia & Present history diarrhea & Present history is pregnant & Present history pregnancy durat \\
Present history ear muffled bil & Present history back pain & Present history blood in stool & Present history common cold sym & Present history sore throat & Present history dysphagia & Present history cough \\
Present history melena & Present history dysuria & Present history snores & Present history nose bleeding & Present history palpitations & Present history treated by phys & Present history flulike symptom \\
Present history mate has notice & Present history body bone muscl & Present history pain in ear & Present history has iron defici & Present history has physiothera & Present history obstipation & Present history pain appears or \\
Present history sputum excretio & Present history chest tightness & Present history chest tightness & Present history two kinds of he & Present history chills & Present history pain after fall & Present history pain in chest o \\
Present history sputum excretio & Present history pleural pain & Present history recent fever & Present history recently finish & Present history ear muffled & Present history fluid out of ea & Present history tympanostomy tu \\
Present history involuntary los & Present history arrives with am & Present history has not taken t & Present history reduced fluid i & Present history reduced food in & Present history worst headache  & Present history pain in joints \\

\hline
\end{tabular}
\caption{\textbf{Tier 1 features.} Part 1 of 2.}
\label{tier01f}

\end{sidewaystable}

\begin{sidewaystable}
\tiny
\centering
\begin{tabular}{llllllllll}
\hline
Present history hemoptysis & Present history pollakisuria & Present history recent surgery & Present history reduced urine o & Present history uneasy & Present history itching & Present history pain in shoulde \\
Present history recent long fli & Present history night sweats & Present history pain in calve a & Present history recent stimulan & Present history is trying to qu & Present history dizziness nauti & Present history bed ridden bc o \\
Present history referred from p & Present history macroscopic hem & Present history inflammation in & Present history throat burn & Present history is using immuno & Present history abdominal pain  & Present history vitals taken af \\
Present history urine incontine & Present history recently diagno & Present history repeated airway & Present history recently diagno & Present history increased o2 ne & Present history bedridden & Present history recently diagno \\
Present history increased leg e & Present history burn in throat & Present history chest pain resp & Present history trouble breathi & Present history feels feverish & Present history urinary stenosi & Present history hard to breath  \\
Present history nocturnal dyspn & Present history unable to use r & Present history confusion & Present history hoarseness & Present history increased sweat & Present history visual field ab & Present history increased clums \\
Present history symptoms have r & Present history lower extremiti & Present history trauma & Present history unlike self acc & Present history hemi symptoms & Present history cough at night & Present history pain caused by  \\
Present history back pain thora & Present history back pain lumbo & Present history neck pain & Present history pain in single  & Present history pain in groin & Present history pain reduction  & Present history saddle numbness \\
Present history morning stiffne & Present history leg length disc & Present history fecal incontine & Present history pain reduction  & Present history unable to work  & Present history pain in buttock & Present history pain increases  \\
Family history migraine & Family history hypertension & Family history heart disease & Family history hemochromatosis & Family history multiple scleros & Family history stroke & Family history of brain tumour \\
Family history of diabetes mell & Family history of deep venous t & Family history of brain aneurys & Family history of lower back di & Pain character pulsating & Pain pain killers work well & Pain onset \\
Pain vas value & Pain stability & Pain character heavy & Pain character tension & Pain character pressure & Pain character sting & Pain radiation neck \\
Pain disturbs sleep & Pain radiation teeth & Pain over maxillary sinuses & Pain vas variable & Pain worsens or gets better wit & Pain radiation to left arm & Pain radiation to jaw \\
Pain radiation to right arm & Pain vas worst value & Pain radiation to back & Pain appears or worsens on vals & Pain changes with food intake & Pain appears or worsens when co & Pain over frontal sinuses \\
Pain appears or worsens with po & Pain appears or worsens on layi & Pain location thorax back & Pain character electrical & Pain appears or worsens when st & Pain appears or worsens when si & Symptom start a few weeks ago \\
Symptom duration 24 hrs or more & Symptom frequency a few times p & Symptom start a few days & Symptom duration one hour or le & Symptom frequency a few times p & Symptom start a few months & Symptom trigger \\
Symptom localisation on the rig & Symptom duration a few hours & Symptom start a year or longer & Symptom localisation on the lef & Symptom frequency every day & Symptom frequency a few times p & Symptom frequency a few times a \\
Symptom start a few hours & Symptom duration is variable & Symptom frequency is variable & Symptom localisation goes betwe & Symptom duration a few minutes & Symptom nsaids work well & Symptom duration a few seconds \\
Symptom start a specific date & Main complaint headache & Main complaint prescription ren & Main complaint nose bleeding & Main complaint visual disturban & Main complaint allergy & Main complaint dizziness \\
Main complaint multiple problem & Main complaint syncope & Main complaint numbness in head & Main complaint back pain & Main complaint pain in knee & Main complaint nausea & Main complaint common cold symp \\
Main complaint aphasia & Main complaint high blood press & Main complaint malaise & Main complaint pain around sing & Main complaint vomiting & Main complaint outside lesion o & Main complaint abdominal pain \\
Main complaint chest pain & Main complaint numbness in limb & Main complaint dyspnea & Main complaint physiotherapy re & Main complaint depression and o & Main complaint car accident & Main complaint shoulder and bac \\
Main complaint shoulder problem & Main complaint pain in joints & Main complaint certificate & Main complaint referral to spec & Main complaint constipation & Main complaint needs prescripti & Main complaint is pregnant \\
Main complaint cough & Main complaint re assessment & Main complaint resp. symp & Main complaint fever & Main complaint pharyngitis & Main complaint pain in chest or & Main complaint pain in ear \\
Main complaint maxillary skin i & Main complaint body bone muscle & Main complaint external tumour  & Main complaint trouble breathin & Main complaint sputum excretion & Main complaint sputum excretion & Main complaint chest tightness \\
Main complaint pleural pain & Main complaint chills & Main complaint impared consciou & Main complaint dysuria & Main complaint migraine & Main complaint palpitations or  & Main complaint asthma exacerbat \\
Main complaint nasal congestion & Main complaint limb reduced for & Main complaint face reduced for & Main complaint feeling unbalanc & Main complaint slurry speech & Main complaint pain in hip & Main complaint pain in lower ex \\
Main complaint pain in buttock  & Cough disturbs sleep & Cough accompanying abdominal pa & Cough barking & Heart rate value & Heart rate value self measureme & Heart rate left side value \\
Heart rate value self measureme & Respiratory frequency value & Oxygen saturation value & Temperature value & Temperature at home value & Blood pressure value & Blood pressure value self measu \\

\hline
\end{tabular}
\caption{\textbf{Tier 1 features.} Part 2 of 2.}
\label{tier0f2}

\end{sidewaystable}

\begin{sidewaystable}
\tiny
\centering
\begin{tabular}{llllllllll}
\hline
Examination lung auscultation a & Examination proprioception abno & Examination is obese & Examination palpable neck lymph & Examination heart auscultation  & Examination systolic heart murm & Examination systolic heart murm \\
Examination abnormal or absent  & Examination abnormal neurologic & Examination abnormal neurologic & Examination pronator drift & Examination positive babinsky & Examination finger nose test ab & Examination rhomberg abnormal \\
Examination abnormal heel to to & Examination abnormal gait & Examination abnormal or asymmet & Examination abnormal neurologic & Examination abnormal sensation  & Examination neurological reflex & Examination is blood pressure e \\
Examination abnormal abdominal  & Examination pupils abnormal & Examination neck stiffness & Examination generally sick look & Examination spontant nystagmus & Examination disturbed eye movem & Examination dix hallpike positi \\
Examination pain with sinus pal & Examination occipital muscles p & Examination slurry speech & Examination is line walking abn & Examination vitals are abnormal & Examination edema & Examination audible carotis bru \\
Examination abnormal or reduced & Examination abnormal force uppe & Examination abnormal force lowe & Examination shoulder muscles pa & Examination restricted neck mov & Examination is overweight & Examination nystagmus \\
Examination abnormal or asymmet & Examination lung auscultation c & Examination abnormal sensation  & Examination abnormal sensation  & Examination abdomen epigastrium & Examination abdomen llq pain on & Examination abdomen rlq pain on \\
Examination lung auscultation w & Examination lung auscultation r & Examination mouth throat abnorm & Examination reflexes patella ab & Examination abnormal or asymmet & Examination ataxia & Examination spurlings test posi \\
Examination lasegue positive si & Examination heart rate irregula & Examination grasset test abnorm & Examination lymph nodes palpabl & Examination otoscopy abnormal b & Examination dysdiadochokinesia  & Examination ram normal \\
Examination pain on scm palpati & Examination fundoscopy abnormal & Examination visual field abnorm & Examination renal pain on percu & Examination abdomen ruq pain on & Examination abdomen pain on rel & Examination otoscopy cerumen bi \\
Examination weak to see & Examination reflexes achilles a & Examination reflexes triceps ab & Examination reflexes biceps abn & Examination thyroid palpation a & Examination costal intercostal  & Examination tonsils enlarged \\
Examination tonsils pus & Examination lumbosacral pain on & Examination face reduced force & Examination language understand & Examination reflexes brachiorad & Examination clonus & Examination rash on body \\
Examination pain on palpation b & Examination otoscopy redness in & Examination pain or no pulse on & Examination pain on palpation p & Examination capillary refill ti & Examination visible petechiae & Examination lung auscultation c \\
Examination lung auscultation c & Examination distal vascular sta & Examination lung auscultation p & Examination otoscopy visible ef & Examination tymp tube not in pl & Examination tymp tube not in pl & Examination otoscopy pus in ear \\
Examination lung auscultation r & Examination trismus & Examination lung ausculation ob & Examination neck venous stasis & Examination abdomen luq pain on & Examination struggles with brea & Examination otoscopy tympanic m \\
Examination abdomen suprapubic  & Examination lung auscultation c & Examination lung auscultation c & Examination abdomen murphys sig & Examination systolic heart murm & Examination o2 value & Examination venous stasis derma \\
Examination skin pallor & Examination tonsils cryptic & Examination otoscopy visible va & Examination otoscopy tymp membr & Examination lung auscultation c & Examination calves pain on palp & Examination tympanic membrane r \\
Examination stridor & Examination using abdominal mus & Examination otoscopy tympanic m & Examination otoscopy visible ef & Examination fontanella abnormal & Examination otoscopy tymp membr & Examination otoscopy tymp membr \\
Examination tympanic membrane r & Examination otoscopy tympanic m & Examination central cyanosis & Examination abdomen diffuse pai & Examination lung auscultation m & Examination calves redness or i & Examination pitting edema lower \\
Examination tympanic membrane r & Examination nose alae flutter & Examination lung deafness on pe & Examination lymph nodes palpabl & Examination pus in eyes bilat & Examination cold extremeties by & Examination mucous membranes dr \\
Examination abdomen visible her & Examination intestinal sounds a & Examination neglect present & Examination sbs value & Examination soft gum abnormal a & Examination lumbo sacral pain o & Examination signs of scoliosis \\
Examination hip reduced range o & Examination pain on palpation t & Examination restricted movement & Examination trouble walking on  & Examination signs of kyphosis & Examination signs of abnormal l & Examination pain on palpation g \\

\hline
\end{tabular}
\caption{\textbf{Tier 2 features.} This tier also includes the previous tier's features.}
\label{tier1f}

\end{sidewaystable}

\begin{sidewaystable}
\tiny
\centering
\begin{tabular}{llllllllll}
\hline
Blood tests tnt value & Blood creatinine value & Blood alat value & Blood total cholesterol value & Blood hdl value & Blood pressure left upper arm v & Blood mcv value \\
Blood tsh value & Blood wbc value & Blood neutrophils value & Blood tests tnt 2 value & Blood d dimer value & Blood bnp value & Blood astrup abormal \\
Blood inr value & Diagnostics blood tests abnorma & Diagnostics blood tests tnt ele & Diagnostics blood status abnorm & Diagnostics blood tests d dimer & Diagnostics blood glucose value & Diagnostics blood esr value \\

\hline
\end{tabular}

\caption{\textbf{Tier 3 features.} This tier also includes the two previous tiers' features.}
\label{tier2f}
\end{sidewaystable}

\end{document}